\documentclass[10pt,twocolumn,letterpaper]{article}

\usepackage{cvpr}
\usepackage{times}
\usepackage{epsfig}
\usepackage{graphicx}
\usepackage{acronym}
\usepackage{amsmath}
\usepackage{amssymb}
\usepackage{tikz}
\usepackage{multirow}
\usepackage{pgfplots}
\usepackage{booktabs}
\usepackage{bbm}
\usepackage{subfig}
\usepackage{dcolumn}
\newcolumntype{d}[1]{D{.}{.}{#1}}
\makeatletter
\newcolumntype{B}[3]{>{\boldmath\DC@{#1}{#2}{#3}}c<{\DC@end}}
\makeatother

\usepackage[pagebackref=true,breaklinks=true,letterpaper=true,colorlinks,bookmarks=false]{hyperref}

\cvprfinalcopy 

\def\cvprPaperID{2555} 

\ifcvprfinal\fi
\begin{document}
	

\acrodef{lrf}[LRF]{local reference frame}
\acrodef{lra}[LRA]{local reference axis}
\acrodef{detmcd}[det-MCD]{deterministic minimum covariance determinant}
\acrodef{tls}[TLS]{terrestrial laser scanning}
\acrodef{mlp}[MLP]{multilayer perceptron}
\acrodef{relu}[ReLU]{rectified linear unit}
\acrodef{nn}[NN]{neural network}
\acrodef{cnn}[CNN]{convolutional neural network}
\acrodef{fc}[FC]{fully connected}
\acrodef{sgd}[SGD]{stochastic gradient descent}
\acrodef{prc}[PRC]{precision-recall curve}
\acrodef{auc}[AUC] {Area under the Curve}
\acrodef{sdv}[SDV] {smoothed density value}
\newcommand{\todo}[1]{\hl{\textit{Tbd: #1}}}
\newcommand{\secPref}{Sect. }
\newcommand{\figPref}{Fig. }
\newcommand{\tabPref}{Table }
\newcommand{\rhl}[1]{\textcolor{red}{\hl{#1}}}


\title{The Perfect Match: 3D Point Cloud Matching with Smoothed Densities}

\author{Zan Gojcic \qquad Caifa Zhou \qquad Jan D. Wegner \qquad Andreas Wieser \\ \\
	ETH Zurich \\
	{\tt\small \{firstname.lastname@geod.baug.ethz.ch\}}}


\newcommand\vv[1]{%
	\begin{tikzpicture}[baseline=(arg.base)]
	\node[inner xsep=0pt] (arg) {$#1$};
	\draw[line cap=round,line width=0.45,->,shorten >= 0.2pt, shorten <= 0.7pt] (arg.north west) -- (arg.north east);
	\end{tikzpicture}%
}

\maketitle
\thispagestyle{empty}

\begin{abstract}
	We propose 3DSmoothNet, a full workflow to match 3D point clouds with a siamese deep learning architecture and fully convolutional layers using a voxelized \acf{sdv} representation. The latter is computed per interest point and aligned to the ~\acf{lrf} to achieve rotation invariance. Our compact, learned, rotation invariant 3D point cloud descriptor achieves $94.9\%$ average recall on the \textit{3DMatch} benchmark data set~\cite{zeng20163dmatch}, outperforming the state-of-the-art by more than $20$ percent points with only 32 output dimensions. This very low output dimension allows for near real-time correspondence search with 0.1 ms per feature point on a standard PC. Our approach is sensor- and scene-agnostic because of~\acs{sdv},~\acs{lrf} and learning highly descriptive features with fully convolutional layers. We show that 3DSmoothNet trained only on RGB-D indoor scenes of buildings achieves $79.0\%$ average recall on laser scans of outdoor vegetation, more than double the performance of our closest, learning-based competitors~\cite{zeng20163dmatch,khoury2017CGF,deng2018ppfnet,Deng2018PPFFoldNetUL}. Code, data and pre-trained models are available online at \url{https://github.com/zgojcic/3DSmoothNet}.
	
\end{abstract}


\section{Introduction}\label{sec:intro}

3D point cloud matching is necessary to combine multiple overlapping scans of a scene (e.g., acquired using an RGB-D sensor or a laser scanner) into a single representation for further processing like 3D reconstruction or semantic segmentation. Individual parts of the scene are usually captured from different viewpoints with a relatively low overlap. A prerequisite for further processing is thus aligning these individual point cloud fragments in a common coordinate system, to obtain one large point cloud of the complete scene.

Although some works aim to register 3D point clouds based on geometric constraints (e.g.,~\cite{rabbani2007,zeisl2013,theiler2015}), most approaches match corresponding 3D feature descriptors that are custom-tailored for 3D point clouds and usually describe point neighborhoods with histograms of point distributions or local surface normals (e.g.,~\cite{johnson1999,flint2007,rusu2009FPFH,tombari2010SHOT}). 
\begin{figure}[!t]
	\begin{center}
		\includegraphics[width=0.92\linewidth]{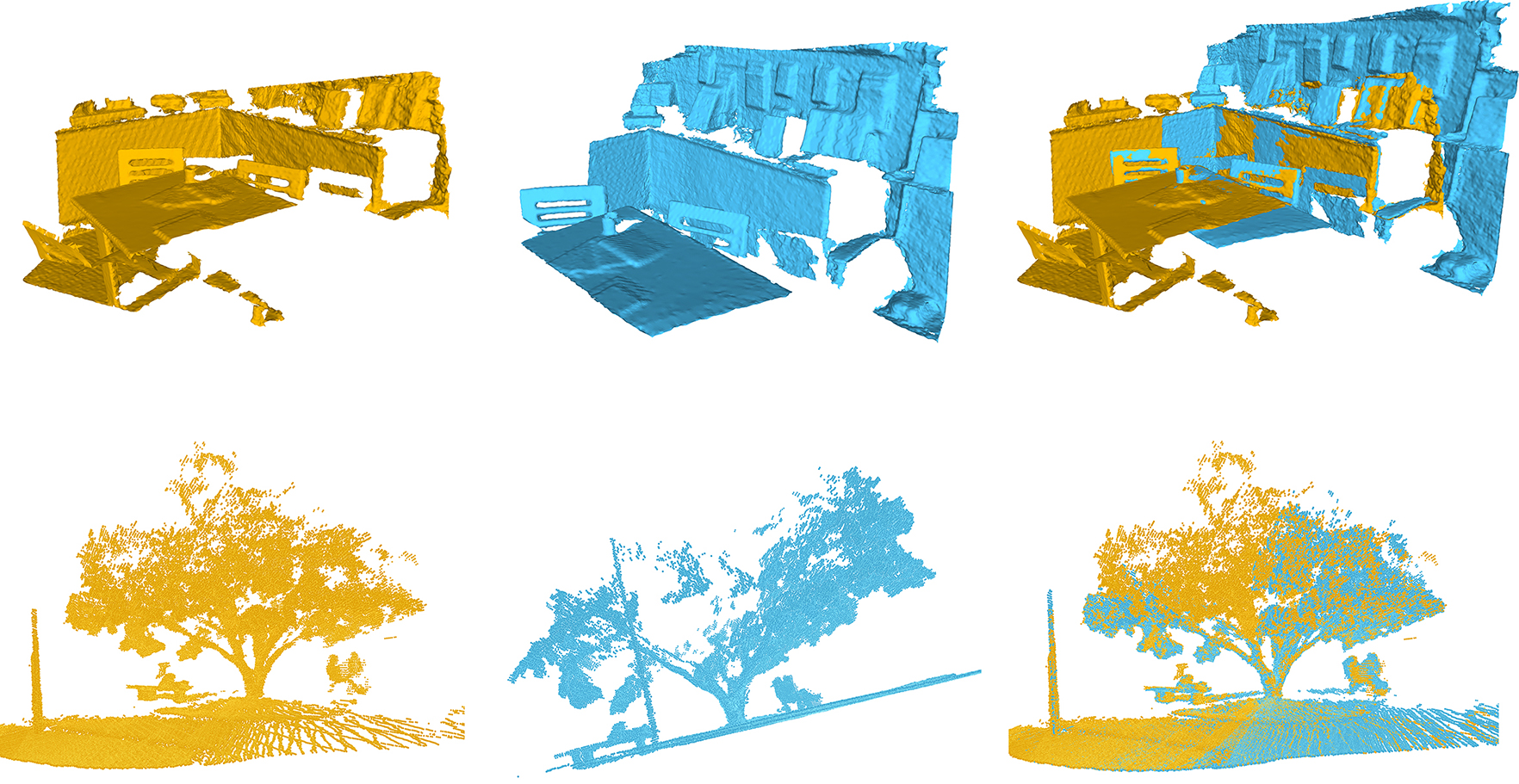}
	\end{center}
	\caption{\textbf{3DSmoothNet generalization ability:} our descriptor, trained solely on indoor scenes (top) can seamlessly generalize to outdoor scenes (bottom).}
	\label{fig:teaser_photo}
	\vspace{-2ex}
\end{figure}
%
%
Since the comeback of deep learning, research on 3D local descriptors has followed the general trend in the vision community and shifted towards learning-based approaches and more specifically deep neural networks~\cite{zeng20163dmatch,khoury2017CGF,deng2018ppfnet,yew20183dfeatnet,Deng2018PPFFoldNetUL}. Although the field has seen significant progress in the last three years, most learned 3D feature descriptors are either not rotation invariant~\cite{zeng20163dmatch,deng2018ppfnet,yew20183dfeatnet}, need very high output dimensions to be successful~\cite{zeng20163dmatch,Deng2018PPFFoldNetUL} or can hardly generalize to new domains~\cite{zeng20163dmatch, khoury2017CGF}.      
In this paper, we propose 3DSmoothNet, a deep learning approach for 3D point cloud matching, which has low output dimension (16 or 32) for very fast correspondence search, high descriptiveness (outperforms all state-of-the-art approaches by more than $20$ percent points), is rotation invariant, and does generalize across sensor modalities and from indoor scenes of buildings to natural outdoor scenes. 
%
\vspace{-3ex}
\paragraph{Contributions} We propose a new compact learned local feature descriptors for 3D point cloud matching that is efficient to compute and outperforms all existing methods significantly. A major technical novelty of our paper is the~\acf{sdv} voxelization as a new input data representation that is amenable to fully convolutional layers of standard deep learning libraries. The gain of~\acs{sdv} is twofold. On the one hand, it reduces the sparsity of the input voxel grid, which enables better gradient flow during backpropagation, while reducing the boundary effects, as well as smoothing out small miss-alignments due to errors in the estimation of the ~\acf{lrf}. On the other hand, we assume that it explicitly models the smoothing that deep networks typically learn in the first layers, thus saving network capacity for learning highly descriptive features. Second, we present a Siamese network architecture with fully convolutional layers that learns a very compact, rotation invariant 3D local feature descriptor. This approach generates low-dimensional, highly descriptive features that generalize across different sensor modalities and from indoor to outdoor scenes. Moreover, we demonstrate that our low-dimensional feature descriptor (only 16 or 32 output dimensions) greatly speeds up correspondence search, which allows real-time applications.
%

\vspace{-1ex}
\section{Related Work}\label{related}
This section reviews advances in 3D local feature descriptors, starting from the early hand-crafted feature descriptors and progressing to the more recent approaches that apply deep learning.
\vspace{-2ex}
\paragraph{Hand-crafted 3D Local Descriptors} Pioneer works on hand-crafted 3D local feature descriptors were usually inspired by their 2D counterparts. Two basic strategies exist depending on how rotation invariance is established. Many approaches including SHOT~\cite{tombari2010SHOT}, RoPS~\cite{Guo2013RoPS}, USC~\cite{tombari2010USC} and TOLDI~\cite{yang2017toldi} try to first estimate a unique \acf{lrf}, which is typically based on the eigenvalue decomposition of the sample covariance matrix of the points in neighborhood of the interest point. This \acs{lrf} is then used to transform the local neighborhood of the interest point to its canonical representation in which the geometric peculiarities, e.g. orientation of the normal vectors or local point density are analyzed. On the other hand, several approaches~\cite{rusu2008PFH, rusu2009FPFH, ilic2015PFH} resort to a \acs{lrf}-free representation based on intrinsically invariant features (e.g., point pair features). Despite significant progress, hand-crafted 3D local descriptors never reached the performance of hand-crafted 2D descriptors. In fact, they still fail to handle point cloud resolution changes, noisy data, occlusions and clutter~\cite{Guo2016descriptorEval}. 
\begin{figure*}[!ht]
	\begin{center}
		\includegraphics[width=0.97\linewidth]{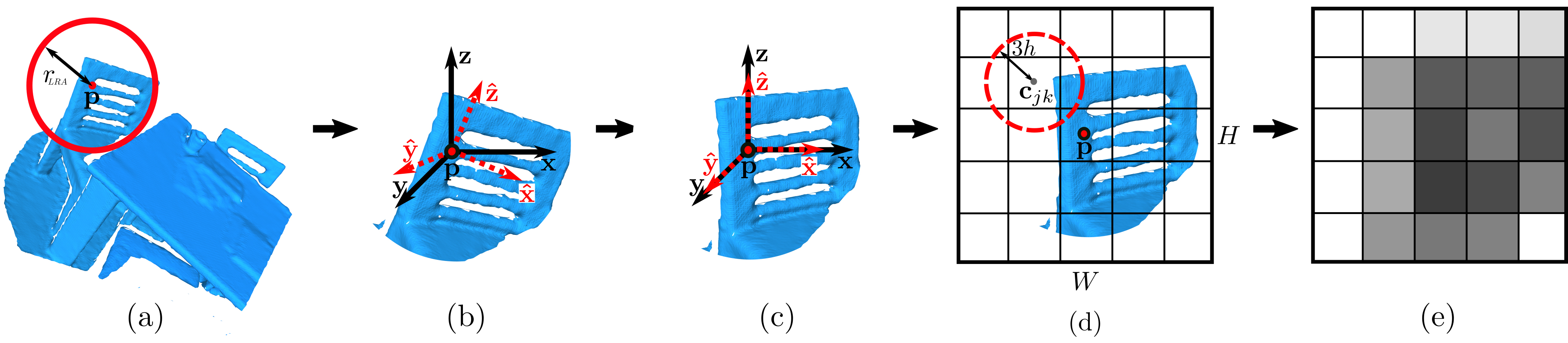}
	\end{center}
	\vspace{-3ex}
	\caption{\textbf{Input parameterization:} (a) We extract the spherical support $\mathcal{S}$ of the interest point $\mathbf{p}$, which is used (b) to estimate a unique \acs{lrf}. (c) Each data cube is transformed to its canonical representation and (d) voxelized using a Gaussian smoothing kernel. (e) The normalized 3D \acs{sdv} voxel grid is used as input to our siamese 3DSmoothNet architecture. Note that (d) and (e) show 2D slices of 3D cubes.}
	\label{fig:input_parametrization}
	\vspace{-1ex}
\end{figure*}
\vspace{-2ex}
\paragraph{Learned 3D Local Descriptors} The success of deep-learning methods in image processing also inspired various approaches for learning geometric representations of 3D data. Due to the sparse and unstructured nature of raw point clouds, several parallel tracks regarding the representation of the input data have emerged. 

One idea is projecting 3D point clouds to images and then inferring local feature descriptors by drawing from the rich library of well-established 2D CNNs developed for image interpretation. For example,~\cite{elbaz2017LORAX} project 3D point clouds to depth maps and extract features using an auto-encoder. ~\cite{huang2018MultiView} use a 2D CNN to combine the rendered views of feature points at multiple scales into a single local feature descriptor. Another possibility are dense 3D voxel grids either in the form of binary occupancy grids~\cite{maturana2015voxnet,wu20153dShape, qi2016volumetric} or an alternative encoding~\cite{zeng20163dmatch,Yarotsky17GeoFeatures}. For example, 3DMatch~\cite{zeng20163dmatch}, one of the pioneer works in learning 3D local descriptors, uses a volumetric grid of truncated distance functions to represent the raw point clouds in a structured manner. Another option is estimating the \acs{lrf} (or a local reference axis) for extracting canonical, high-dimensional, but hand-crafted features and using a neural network solely for a dimensionality reduction. Even-though these methods ~\cite{khoury2017CGF,gojcic2018learned} manage to learn a non-linear embedding that outperforms the initial representations, their performance is still bounded by the descriptiveness of the initial hand-crafted features.

PointNet~\cite{qi2017pointnet} and PointNet++~\cite{qi2017pointnet++} are seminal works that introduced a new paradigm by directly working on raw unstructured point-clouds. They have shown that a permutation invariance of the network, which is important for learning on unordered sets, can be accomplished by using symmetric functions. Albeit, successful in segmentation and classification tasks, they do not manage to encapsulate the local geometric information in a satisfactory manner, largely because they are unable to use convolutional layers in their network design~\cite{deng2018ppfnet}. Nevertheless, PointNet offers a base for PPFNet~\cite{deng2018ppfnet}, which augments raw point coordinates with point-pair features and incorporates global context during learning to improve the feature representation. However, PPFNet is not fully rotation invariant. PPF-FoldNet addresses the rotation invariance problem of PPFNet by solely using point-pair features as input. It is based on the architectures of the PointNet~\cite{qi2017pointnet} and FoldingNet~\cite{yaoqing2018FoldingNet} and is trained in a self-supervised manner. 
The recent work of \cite{yew20183dfeatnet} is based on PointNet, too, but deviates from the common approach of learning only the feature descriptor. It follows the idea of~\cite{yi2016lift} in trying to fuse the learning of the keypoint detector and descriptor in a single network in a weakly-supervised way using  GPS/INS tagged 3D point clouds. \cite{yew20183dfeatnet} does not achieve rotation invariance of the descriptor and is limited to smaller point cloud sizes due to using PointNet as a backbone. 

Arguably, training a network directly from raw point clouds fulfills the end-to-end learning paradigm. On the downside, it does significantly hamper the use of convolutional layers, which are crucial to fully capture local geometry. 
We thus resort to a hybrid strategy that, first, transforms point neighborhoods into \acs{lrf}s, second, encodes unstructured 3D point clouds as \acs{sdv} grids amenable to convolutional layers and third, learns descriptive features with a siamese CNN. This strategy does not only establish rotation invariance but also allows good performance with low output dimensions, which speeds up correspondence search.

\section{Method}\label{method}
\begin{figure*}[!ht]
	\begin{center}
		\includegraphics[width=0.94\linewidth]{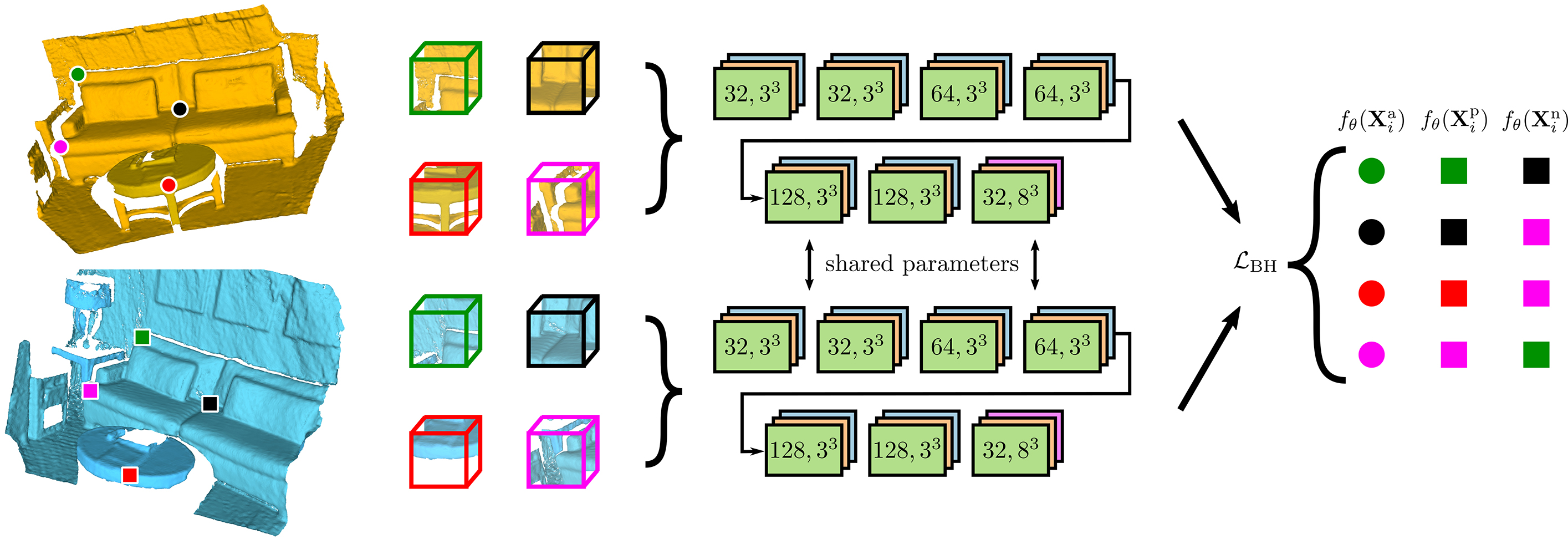}
	\end{center}
	\vspace{-2ex}
	\caption{\textbf{3DSmoothNet network architecture:} We extract interest points in the overlapping region of two fragments. The cubic patches (bounding box is color coded to the interest points), centered at the interest point and aligned with the estimate \acs{lrf} are converted to the \acs{sdv} voxel grid and fed to the network. 3DSmoothNet consists of convolutional (green rectangles with number of filters and filter size respectively), batch-normalization (orange), ReLU activation function (blue) and an $l2$-normalization (magenta) layer. Both branches share all the parameters. The anchor $f_\theta(\mathbf{X}^\text{a})$, positive $f_\theta(\mathbf{X}^\text{p})$ and negative  $f_\theta(\mathbf{X}^\text{n})$ arguments of the batch hard loss are color coded according to the interest points. Negative examples are sampled on the fly from all the positive examples of the mini-batch (denoted with the four voxel grids).}
	\label{fig:network_architecture}
\end{figure*}
In a nutshell, our workflow is as follows (Fig.~\ref{fig:input_parametrization} \&~\ref{fig:network_architecture}): (i) given two raw point clouds, (ii) compute the \acs{lrf} of the spherical neighborhood around the randomly selected interest points, (iii) transform the neighborhoods to their canonical representations, (iv) voxelize them with the help of Gaussian smoothing, (v) infer the per point local feature descriptors using 3DSmoothNet and, for example, use them as input to a RANSAC-based robust point cloud registration pipeline.

More formally, consider two overlapping point cloud sets $\mathcal{P}$ and $\mathcal{Q}$ represented in matrix form as $\mathbf{P} \in \mathbb{R}^{n \times 3}$ and $\mathbf{Q} \in \mathbb{R}^{m \times 3}$. Let $\mathbf{(P)}_i =: \mathbf{p}_i$ represent the coordinate vector of an individual point of point cloud $\mathcal{P}$ located in the overlapping region. A bijective function maps point $\mathbf{p}_i$ to its corresponding (but initially unknown) point $(\mathbf{Q})_j =: \mathbf{q}_j$ in the second point cloud. Under the assumption of a static scene and rigid point clouds (and neglecting noise and differing point cloud resolutions), this bijective function can be described with the transformation parameters of the congruent transformation
\begin{equation}
\mathbf{q}_j = \mathbf{R}\mathbf{p}_i + \mathbf{t},
\end{equation}
where $\mathbf{R} \in SO(3)$ denotes the rotation matrix and $\mathbf{t} \in \mathbb{R}^{3}$ the translation vector. With point subsets $\mathbf{P}^c$ and $\mathbf{Q}^c$ for which correspondences exist, the mapping function can be written as
\begin{equation}
\label{eq:transformation_setting}
\mathbf{Q}^c = \mathbf{K}\mathbf{P}^c\mathbf{R}^{T} + \mathbf{1}\otimes\mathbf{t}^{T},
\end{equation}
where $\mathbf{K} \in \mathbb{P}^{|\mathbf{Q'}|}$ denotes a permutation matrix whose entries $k_{ij} = 1$ if $\mathbf{p}_i$ corresponds to $\mathbf{q}_j$ and $0$ otherwise and $\mathbf{1}$ is a vector of ones. In our setting both, the permutation matrix $\mathbf{K}$ and the transformation parameters $\mathbf{R}$ and $\mathbf{t}$ are unknown initially.  Simultaneously solving for all is hard as the problem is non-convex with binary entries in $\mathbf{K}$~\cite{li20073d}. However, if we find a way to determine $\mathbf{K}$, the estimation of the transformation parameters is straightforward. 
This boils down to learning a function that maps point $\mathbf{p}_i$ to a higher dimensional feature space in which we can determine its corresponding point $\mathbf{q}_j$. Once we have established correspondence, we can solve for $\mathbf{R}$ and $\mathbf{t}$. 
Computing a rich feature representation across the neighborhood of $\mathbf{p}_i$ ensures robustness against noise and facilitates high descriptiveness. Our main objective is a fully rotation invariant local feature descriptor that generalizes well across a large variety of scene layouts and point cloud matching settings. We choose a data-driven approach for this task and learn a compact local feature descriptor from raw point clouds.
\subsection{Input parameterization}\label{sec:input_param}
\begin{figure}[!t]
	\begin{center}
		\includegraphics[width=\linewidth]{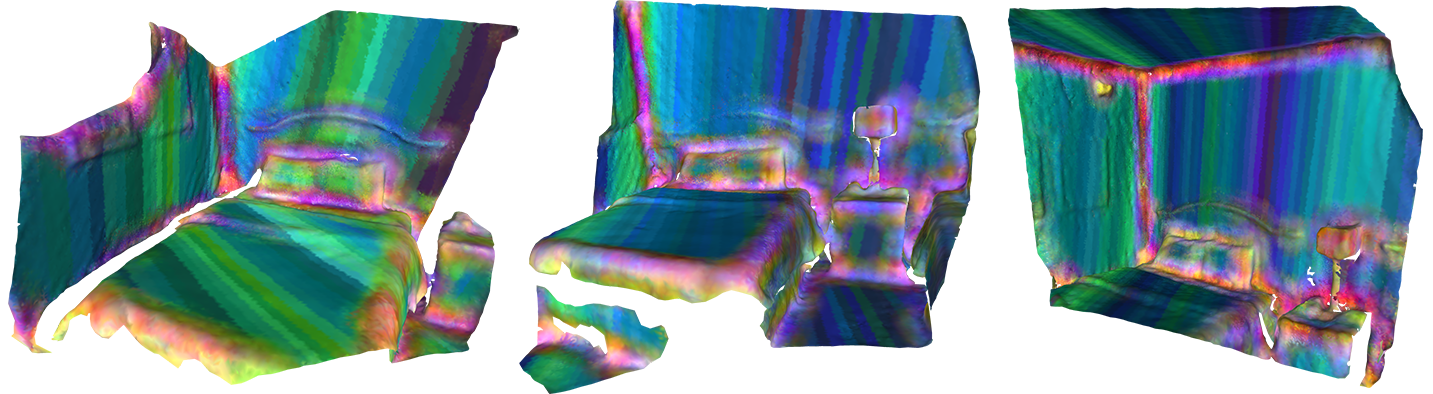}
	\end{center}
	\vspace{-2ex}
	\caption{\textbf{3DSmoothNet descriptors are geometrically informed:} Embedding in 3D space with PCA (first three components RGB color-coded). Planar regions lie in the blue-green, edges and corners in the orange-pink and spherical surfaces in the yellow color spectrum. }
	\label{fig:pca_representation}
\end{figure}

A core requirement for a generally applicable local feature descriptor is its invariance under isometry of the Euclidian space. Since achieving the rotation invariance in practice is non-trivial, several recent works~\cite{zeng20163dmatch,deng2018ppfnet,yew20183dfeatnet} choose to ignore it and thus do not generalize to rigidly transformed scenes~\cite{Deng2018PPFFoldNetUL}. One strategy to make a feature descriptor rotation invariant is regressing the canonical orientation of a local 3D patch around a point as an integral part of a deep neural network~\cite{qi2017pointnet,yew20183dfeatnet} inspired by recent work in 2D image processing~\cite{jaderberg2015spatial, yi2016learning}. However, ~\cite{deng2018ppfnet, esteves2018learning} find that this strategy often fails for 3D point clouds. We therefore choose a different approach and explicitly estimate \acs{lrf}s by adapting the method of~\cite{yang2017toldi}. An overview of our method is shown in Fig.~\ref{fig:input_parametrization} and is described in the following.
\paragraph{Local reference frame} Given a point $\mathbf{p}$ in point cloud $\mathcal{P}$, we select its local spherical support $\mathcal{S}\subset\mathcal{P}$ such that $\mathcal{S} = \{\mathbf{p}_i:||\mathbf{p}_i-\mathbf{p}||_2 \leq r_{LRF}\}$ where $r_{LRF}$ denotes the radius of the local neighborhood used for estimating the \acs{lrf}. In contrast to \cite{yang2017toldi}, where only points within the distance $\frac{1}{3}r_{LRF}$ are used, we approximate the sample covariance matrix $\mathbf{\tilde{\Sigma}}_\mathcal{S}$ using all the points $\mathbf{p}_i \in \mathcal{S}$. Moreover, we replace the centroid with the interest point $\mathbf{p}$ to reduce computational complexity. We compute the \acs{lrf} via the eigendecomposition of $\mathbf{\tilde{\Sigma}_\mathcal{S}}$:
\begin{equation}
\mathbf{\tilde{\Sigma}}_\mathcal{S} = \frac{1}{|\mathcal{S}|} \sum_{\mathbf{p}_i \in \mathcal{S}}(\mathbf{p}_i-\mathbf{p})(\mathbf{p}_i-\mathbf{p})^{\rm T}
\end{equation}
We select the z-axis $\mathbf{\hat{z}}_\mathbf{p}$ to be collinear to the estimated normal vector $\hat{\mathbf{n}}_\mathbf{p}$ obtained as the eigenvector corresponding to the smallest eigenvalue of $\mathbf{\tilde{\Sigma}}_\mathcal{S}$. We solve for the sign ambiguity of the normal vector $\mathbf{\hat{z}}_\mathbf{p}$ by
\begin{equation}
\mathbf{\hat{z}}_\mathbf{p} =\begin{cases} \:\:\;\hat{\mathbf{n}}_\mathbf{p}, \:\:& \text{if } \sum\limits_{\mathbf{p}_i \in \mathcal{S}} \langle\hat{\mathbf{n}}_\mathbf{p},\vv{\mathbf{p}_i\mathbf{p}}\rangle \geq 0\\
-\hat{\mathbf{n}}_\mathbf{p}, \:\:& \text{otherwise}
\end{cases}
\end{equation}
The x-axis $\mathbf{\hat{x}}_\mathbf{p}$ is computed as the weighted vector sum
\begin{equation}
\mathbf{\hat{x}}_\mathbf{p} = \frac{1}{ ||\sum\limits_{\mathbf{p}_i \in \mathcal{S}} \alpha_{i} \beta_{i} \mathbf{v}_i||_2} \sum\limits_{\mathbf{p}_i \in \mathcal{S}} \alpha_{i} \beta_{i} \mathbf{v}_i
\end{equation}
where $\mathbf{v}_i =\vv{\mathbf{p}\mathbf{p}_i} - \langle\vv{\mathbf{p}\mathbf{p}_i}
,\mathbf{\hat{z}}_\mathbf{p} \rangle \mathbf{\hat{z}}_\mathbf{p}$ is the projection of the vector $\vv{\mathbf{p}\mathbf{p}_i}$ to a plane orthogonal to $\mathbf{\hat{z}}_\mathbf{p}$ and $\alpha_{i}$ and $\beta_{i}$ are weights related to the norm and the scalar projection of the vector $\vv{\mathbf{p}\mathbf{p}_i}$ to the vector $\mathbf{\hat{z}}_\mathbf{p}$ computed as
\begin{equation}
\begin{split}
\alpha_{i} &= (r_{LRF} - ||\mathbf{p} - \mathbf{p}_i ||_2 )^2 \\
\beta_{i}  &= \langle \vv{\mathbf{p}\mathbf{p}_i},\mathbf{\hat{z}}_\mathbf{p}\rangle^2
\end{split}
\end{equation}
Intuitively, the weight $\alpha_{i}$ favors points lying close to the interest point thus making the estimation of $\mathbf{\hat{x}}_\mathbf{p}$ more robust against clutter and occlusions. $\beta_{i}$ gives more weight to points with a large scalar projection, which are likely to contribute significant evidence particularly in planar areas~\cite{yang2017toldi}. Finally, the y-axis $\mathbf{\hat{y}}_\mathbf{p}$ completes the left-handed \acs{lrf}  and is computed as  $\mathbf{\hat{y}}_\mathbf{p}  = \mathbf{\hat{x}}_\mathbf{p} \times \mathbf{\hat{z}}_\mathbf{p}$.
\begin{figure*}[!ht]
	\begin{center}
		\includegraphics[width=0.92\linewidth]{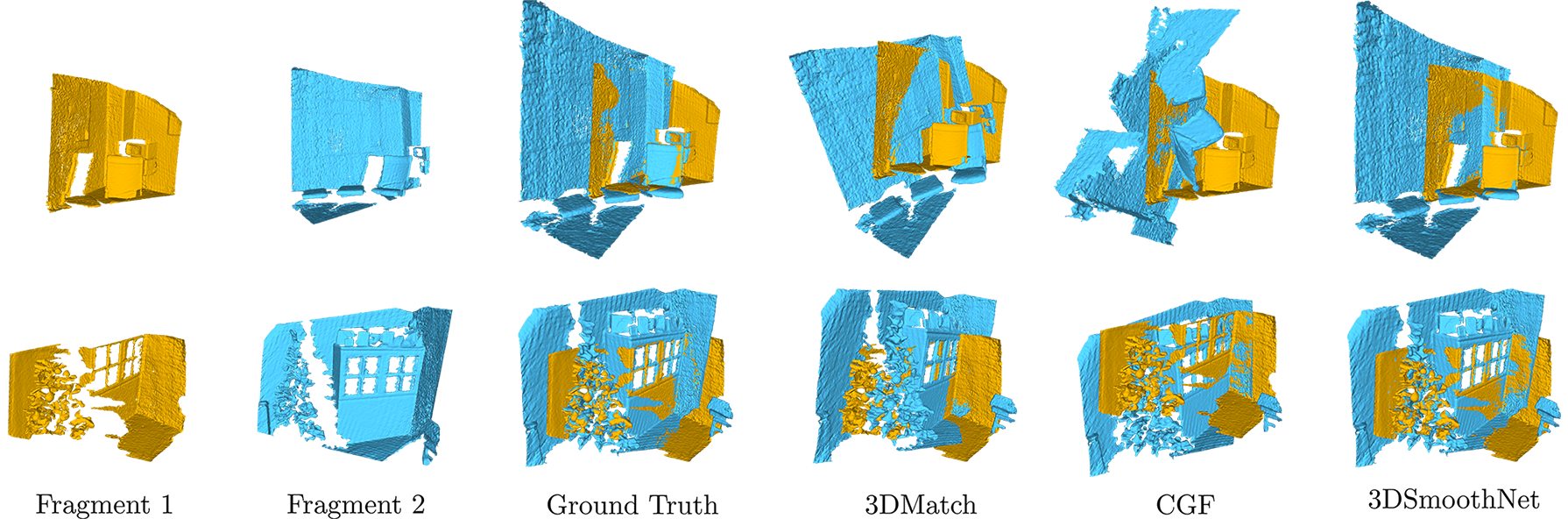}
	\end{center}
	\vspace{-2ex}
	\caption{\textbf{Results on the \textit{3DMatch} data set after RANSAC:} 3DSmoothNet generates reliable correspondences for pairs with low overlap ($32\%$ (top), $48\%$ (bottom)) and predominantly planar regions (top row) or hard cases with vegetation and repetitive geometries (Christmas tree, windows in bottom row).}
	\label{fig:registarion_examples}
\end{figure*}
\paragraph{Smoothed density value (SDV) voxelization}
Once points in the local neighborhood ${\mathbf{p}_i \in \mathcal{S}}$ have been transformed to their canonical representation $\mathbf{p'}_i \in \mathcal{S'}$ (Fig.~\ref{fig:input_parametrization}(c)), we use them to describe the transformed local neighborhood of the interest points. We represent points in a \acs{sdv} voxel grid, centered on the interest point $\mathbf{p'}$ and aligned with the \acs{lrf}. We write the \acs{sdv} voxel grid as a three dimensional matrix $\mathbf{X}^{\text{SDV}} \in \mathbb{R}^{W \times H \times D}$ whose elements $(\mathbf{X}^{\text{SDV}})_{jkl} =:  x_{jkl}$ represent the \acs{sdv} of the corresponding voxel computed using the Gaussian smoothing kernel with bandwidth $h$: 
\begin{equation}
\begin{split}
x_{jkl} = \frac{1}{n_{jkl}}&\sum_{i=1}^{n_{jkl}} \frac{1}{\sqrt{2\pi}h}\exp{\frac{-||\mathbf{c}_{jkl}-\mathbf{p'}_i||_2^2}{2h^2}} \\
&\text{s.t.} \; ||\mathbf{c}_{jkl}-\mathbf{p'}_i||_2 < 3h
\end{split}
\end{equation}
where $n_{jkl}$ denotes the number of points $\mathbf{p'}_i \in \mathcal{S'}$ that lie within the distance $3h$ of the voxel centroid $\mathbf{c}_{jkl}$ (see Fig.~\ref{fig:input_parametrization}(d)). Further, all values of $\mathbf{X}^{\text{SDV}}$ are normalized such that they sum up to $1$ in order to achieve invariance with respect to varying point cloud densities. For ease of notation we omit the superscript SDV in $\mathbf{X}^\text{SDV}$ in all following equations. 
The proposed \acs{sdv} voxel grid representation has several advantages over the traditional binary-occupancy grid \cite{maturana2015voxnet, wu20153dShape}, the truncated distance function \cite{zeng20163dmatch} or hand-crafted feature representations \cite{khoury2017CGF, gojcic2018learned,deng2018ppfnet,Deng2018PPFFoldNetUL}. First, we mitigate the impact of boundary effects and noise of binary-occupancy grids and truncated distance functions by smoothing density values over voxels. Second, compared to the binary occupancy grid we reduce the sparsity of the representation on the fragments of the test part of \textit{3DMatch} data set by more than $30$ percent points (from about $90\%$ to about $57\%$), which enables better gradient flow during backpropagation. Third, the \acs{sdv} representation helps our method to achieve better generalization as we do not overfit exact data cues during training. Finally, in contrast to hand-crafted feature representations, \acs{sdv} voxel grid representation provides input with a geometrically informed structure, which enables us to exploit convolutional layers that are crucial to capture the local geometric characteristics of point clouds (Fig.~\ref{fig:pca_representation}).
\vspace{-2ex}
\paragraph{Network architecture}
Our network architecture (Fig.~\ref{fig:network_architecture}) is loosely inspired by L2Net~\cite{tian2017l2Net}, a state-of-the-art learned local image descriptor. 3DSmoothNet consists of stacked convolutional layers that applies strides of $2$ (instead of max-pooling) in some convolutional layers to down-sample the input \cite{springenberg2015ALLCNN}. All convolutional layers, except the final one, are followed by batch normalization~\cite{ioffe2015batch} and use the ReLU activation function~\cite{nair2010relu}. In our implementation, we follow~\cite{tian2017l2Net} and fix the affine parameters of the batch normalization layer to 1 and 0 and we do not train them during the training of the network. To avoid over-fitting the network, we add dropout regularization~\cite{srivastava2014dropout} with a $0.3$ dropout rate before the last convolutional layer. The output of the last convolutional layer is fed to a batch normalization layer followed by an $l2$ normalization to produce unit length local feature descriptors.
\vspace{-2ex}
\paragraph{Training}\label{sec:CNN_Architecture} We train 3DSmoothNet (Fig.~\ref{fig:network_architecture}) on point cloud fragments from the \textit{3DMatch} data set \cite{zeng20163dmatch}. This is an RGB-D data set consisting of 62 real-world indoor scenes, ranging from offices and hotel rooms to tabletops and restrooms. Point clouds obtained from a pool of data sets~\cite{xiao2013sun3d, shotton2013scene, lai2014unsupervised, valentin2016learning, dai2017bundlefusion} are split into 54 scenes for training and 8 scenes for testing. Each scene is split into several partially overlapping fragments with their ground truth transformation parameters $T$. 

Consider two fragments $\mathcal{F}_i$ and $\mathcal{F}_j$, which have more than $30\%$ overlap. To generate training examples, we start by randomly sampling $300$ anchor points $\mathbf{p}^\text{a}$ from the overlapping region of fragment $\mathcal{F}_i$. After applying the ground truth transformation parameters $T_j()$ the positive sample $\mathbf{p}^\text{p}$ is then represented as the nearest-neighbor $\mathbf{p}^\text{p} =: \text{nn}(\mathbf{p}^\text{a}) \in T_j(\mathcal{F}_j)$, where $\text{nn}()$ denotes the nearest neighbor search in the Euclidean space based on the $l2$ distance. We refrain from pre-sampling the negative examples and instead use the hardest-in-batch method~\cite{hermans2017tripletLoss}  for sampling negative samples on the fly. During training we aim to minimize the soft margin Batch Hard (BH) loss function
\begin{equation}
\begin{split}
\mathcal{L}_\text{BH}(\theta,\mathcal{X}) = \frac{1}{|\mathcal{X}|}\sum_{i=1}^{|\mathcal{X}|} \ln \Bigl(1 +\exp\bigl[ ||f_\theta(\mathbf{X}_i^\text{a})-f_\theta(\mathbf{X}_{i}^\text{p})||_2  \Bigr. \bigr.\\ \Bigl. \bigl.-\min_{\substack{j=1...|\mathcal{X}|\\ j \neq i }} ||f_\theta(\mathbf{X}_i^\text{a})-f_\theta(\mathbf{X}_{j}^\text{p})||_2 \bigr]\Bigr)
\end{split}
\label{eq:batch_hard}
\end{equation}
The BH loss is defined for a mini-batch $\mathcal{X}$, where $\mathbf{X}_{\text{i}}^a$ and $\mathbf{X}_{\text{i}}^p$ represent the \acs{sdv} voxel grids of the anchor and positive input samples, respectively. The negative samples are retrieved as the hardest non-corresponding positive samples in the mini-batch (c.f. Eq.~\ref{eq:batch_hard}). Hardest-in-batch sampling ensures that negative samples are neither too easy (i.e, non-informative) nor exceptionally hard, thus preventing the model to learn normal data associations~\cite{hermans2017tripletLoss}.
\section{Results}\label{experiments}
\paragraph{Implementation details} Our 3DSmoothNet approach is implemented in C++ (input parametrization) using the PCL~\cite{Rusu_ICRA2011_PCL} and in Python (CNN part) using Tensorflow~\cite{tensorflow2015-whitepaper}. During training we extract \acs{sdv} voxel grids of size $W = H = D = 0.3~m$ (corresponding to \cite{zeng20163dmatch}), centered at each interest point and aligned with the \acs{lrf}. We use $r_{\text{LRF}} = \sqrt{3}W$ to extract the spherical support $\mathcal{S}$ and estimate the \acs{lrf}. We obtain the circumscribed sphere of our voxel grid and use the points transformed to the canonical frame to extract the \acs{sdv} voxel grid. We split each \acs{sdv} voxel grid into $16^3$ voxels with an edge $w = \frac{W}{16}$ and use a Gaussian smoothing kernel with an empirically determined optimal width $h = \frac{1.75 w}{2}$. All the parameters wew slected on the validation data set. We train the network with mini-batches of size $256$ and optimize the parameters with the ADAM optimizer \cite{Kingma2015ADAM}, using an initial learning rate of $0.001$ that is exponentially decayed every $5000$ iterations. Weights are initialized orthogonally~\cite{saxe2013orthogonal} with $0.6$ gain, and biases are set to $0.01$. We train the network for $20$ epochs.

We evaluate the performance of 3DSmoothNet for correspondence search on the \textit{3DMatch} data set~\cite{zeng20163dmatch} and compare against the state-of-the-art. In addition, we evaluate its generalization capability to a different sensor modality (laser scans) and different scenes (e.g., forests) on the \textit{Challenging data sets for point cloud registration algorithms} data set~\cite{Pomerleau2012_ETHdataset} denoted as \textit{ETH} data set.
\vspace{-2ex}
\paragraph{Comparison to state-of-the-art} We adopt the commonly used hand-crafted 3D local feature descriptors FPFH~\cite{rusu2009FPFH} (33 dimensions) and SHOT~\cite{tombari2010SHOT} (352 dimensions) as baselines and run implementations provided in PCL~\cite{Rusu_ICRA2011_PCL} for both approaches.
We compare against the current state-of-the-art in learned 3D feature descriptors: 3DMatch~\cite{zeng20163dmatch} (512 dimensions), CGF~\cite{khoury2017CGF} (32 dimensions), PPFNet~\cite{deng2018ppfnet} (64 dimensions), and PPF-FoldNet~\cite{Deng2018PPFFoldNetUL} (512 dimensions). In case of 3DMatch and CGF we use the implementations provided by the authors in combination with the given pre-trained weights. Because source-code of PPFNet and PPF-FoldNet is not publicly available, we report the results presented in the original papers. For all descriptors based on the normal vectors, we ensure a consistent orientation of the normal vectors across the fragments. To allow for a fair evaluation, we use exactly the same interest points (provided by the authors of the data set) for all descriptors. In case of descriptors that are based on spherical neighborhoods, we use a radius that yields a sphere with the same volume as our voxel. All exact parameter settings, further implementation details etc. used for these experiments are available in supplementary material. 
%
\subsection{Evaluation on the 3DMatch data set}
\label{sec:3DMatch_data set}
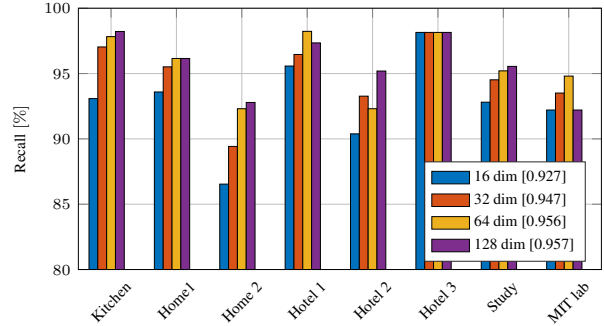
\begin{figure}
	\centering
	\resizebox{0.95\linewidth}{!}{
%
%
\definecolor{mycolor1}{rgb}{0.00000,0.44700,0.74100}%
\definecolor{mycolor2}{rgb}{0.85000,0.32500,0.09800}%
\definecolor{mycolor3}{rgb}{0.92900,0.69400,0.12500}%
\definecolor{mycolor4}{rgb}{0.49400,0.18400,0.55600}%
\begin{tikzpicture}

\begin{axis}[%
width=\columnwidth,
height=0.5\columnwidth,
at={(0\columnwidth,0\columnwidth)},
scale only axis,
log origin=infty,
xmin=0.5,
xmax=8.5,
xtick={1,2,3,4,5,6,7,8},
xticklabels={{Kitchen},{Home1},{Home 2},{Hotel 1},{Hotel 2},{Hotel 3},{Study},{MIT lab}},
xticklabel style={rotate=45,font=\scriptsize},
ymin=80,
ymax=100,
ylabel={Recall $[\%]$},
ylabel near ticks,
yticklabel style={font=\scriptsize},
label style={font=\scriptsize},
axis background/.style={fill=white},
xmajorgrids,
ymajorgrids,
legend style={at={(0.97,0.03)},anchor=south east,legend cell align=left,font=\scriptsize, align=left,draw=white!15!black}
]
\addplot[ybar,bar width=4,bar shift=-8,fill=mycolor1,draw=black,area legend] plot table[row sep=crcr] {%
1	93.083\\
2	93.59\\
3	86.538\\
4	95.575\\
5	90.385\\
6	98.148\\
7	92.808\\
8	92.208\\
};
\addlegendentry{16 dim [0.927]};

\addplot[ybar,bar width=4,bar shift=-4,fill=mycolor2,draw=black,area legend] plot table[row sep=crcr] {%
1	97.036\\
2	95.513\\
3	89.423\\
4	96.46\\
5	93.269\\
6	98.148\\
7	94.52\\
8	93.507\\
};
\addlegendentry{32 dim [0.947]};

\addplot[ybar,bar width=4,bar shift=0,fill=mycolor3,draw=black,area legend] plot table[row sep=crcr] {%
1	97.826\\
2	96.154\\
3	92.308\\
4	98.23\\
5	92.308\\
6	98.148\\
7	95.206\\
8	94.805\\
};
\addlegendentry{64 dim [0.956]};

\addplot[ybar,bar width=4,bar shift=4,fill=mycolor4, draw=black,area legend] plot table[row sep=crcr] {%
1	98.221\\
2	96.154\\
3	92.788\\
4	97.345\\
5	95.192\\
6	98.148\\
7	95.548\\
8	92.208\\
};
\addlegendentry{128 dim [0.957]};

\end{axis}
\end{tikzpicture}%
}
	\caption{\textbf{Recall in relation to 3DSmoothNet output dimensions.} Values in brackets denote average recall over all scenes.}
	\label{fig:dimensionality}
\end{figure}

\paragraph{Setup} The test part of the \textit{3DMatch} data set consists of $8$ indoor scenes split into several partially overlapping fragments. For each fragment, the authors provide indices of $5000$ randomly sampled feature points. We use these feature points for all descriptors. The results of PPFNet and PPF-FoldNet are based on a spherical neighborhood with a diameter of $0.6\text{m}$. Furthermore, due to its memory bottleneck, PPFNet is limited to $2048$ interest points per fragment.
We adopt the evaluation metric of~\cite{deng2018ppfnet} (see supplementary material). It is based on the theoretical analysis of the number of iterations needed by a robust registration pipeline, e.g. RANSAC, to find the correct set of transformation parameters between two fragments. As done in~\cite{deng2018ppfnet}, we set the threshold $\tau_1 = 0.1m$ on the $l2$ distance between corresponding points in the Euclidean space and $\tau_2 = 0.05$ to threshold the inlier ratio of the correspondences at $5\%$.
\vspace{-2ex}
\paragraph{Output dimensionality of 3DSmoothNet} A general goal is achieving the highest matching performance with the lowest output dimensionality (i.e., filter number in the last convolutional layer of 3DSmoothNet) to decrease run-time and to save memory. Thus, we first run trials to find a good compromise between matching performance and efficiency for the 3DSmoothNet descriptors\footnote{Recall that for correspondence search, the brute-force implementation of nearest-neighbor search scales with $\mathcal{O}(DN^2)$, where $D$ denotes the dimension and $N$ the number of data points. The time complexity can be reduced to $\mathcal{O}(DN\log{N})$ using tree-based methods, but still becomes inefficient if $D$ grows large ("curse of dimensionality").}. 
%
We find that the performance of 3DSmoothNet quickly starts to saturate with increasing output dimensions (Fig.~\ref{fig:dimensionality}). There is only marginal improvement (if any) when using more than $64$ dimensions. We thus decide to process all further experiments only for $16$ and $32$ output dimensions of 3DSmoothNet. 
\begin{figure}[!ht]
	\centering
	\resizebox{0.97\linewidth}{!}{
%
%
\definecolor{mycolor1}{rgb}{0.00000,0.44700,0.74100}%
\definecolor{mycolor2}{rgb}{0.85000,0.32500,0.09800}%
\definecolor{mycolor3}{rgb}{0.92900,0.69400,0.12500}%
\definecolor{mycolor4}{rgb}{0.49400,0.18400,0.55600}%
\definecolor{mycolor5}{rgb}{0.46600,0.67400,0.18800}%
\definecolor{mycolor6}{rgb}{0.30100,0.74500,0.93300}%
\begin{tikzpicture}

\begin{axis}[%
width=\columnwidth,
height=0.5\columnwidth,
at={(0\columnwidth,0\columnwidth)},
scale only axis,
scaled x ticks=false,
xmin=0.01,
xmax=0.2,
xlabel={$\tau_2$},
xtick={0,0.04,0.08,0.12,0.16,0.2},
xticklabels={0,0.04,0.08,0.12,0.16,0.2},
ymin=0,
ymax=100,
xticklabel style={font=\footnotesize},
ylabel={Recall $[\%]$},
ylabel near ticks,
yticklabel style={font=\footnotesize},
xticklabel style={font=\footnotesize},
label style={font=\footnotesize},
axis background/.style={fill=white},
xmajorgrids,
ymajorgrids,
legend style={at={(0.02,0.02)},anchor=south west,font=\scriptsize,row sep=-2pt,legend cell align=left,align=left,draw=white!15!black}
]
\addplot [color=mycolor1,line width=1.5pt]
  table[row sep=crcr]{%
0.01	98.66\\
0.02	97.21\\
0.03	96.11\\
0.04	94.37\\
0.05	92.79\\
0.06	90.87\\
0.07	88.73\\
0.08	85.81\\
0.09	83.64\\
0.1	81.69\\
0.11	79.47\\
0.12	78.12\\
0.13	76.22\\
0.14	74.08\\
0.15	71.79\\
0.16	70.44\\
0.17	68.6\\
0.18	65.67\\
0.19	64.26\\
0.2	62.16\\
};
\addlegendentry{Ours (16)};

\addplot [color=mycolor2,line width=1.5pt]
  table[row sep=crcr]{%
0.01	99.05\\
0.02	97.73\\
0.03	97.11\\
0.04	96.12\\
0.05	94.73\\
0.06	93.38\\
0.07	92.68\\
0.08	91.39\\
0.09	90.21\\
0.1	87.99\\
0.11	86.55\\
0.12	84.65\\
0.13	83.56\\
0.14	82.47\\
0.15	81.03\\
0.16	79.27\\
0.17	77.49\\
0.18	75.84\\
0.19	74.26\\
0.2	72.66\\
};
\addlegendentry{Ours (32)};

\addplot [color=mycolor3,line width=1.5pt]
  table[row sep=crcr]{%
0.01	92.93\\
0.02	83.89\\
0.03	74.05\\
0.04	64.5\\
0.05	58.19\\
0.06	51.28\\
0.07	45.37\\
0.08	40.8\\
0.09	36.45\\
0.1	33.38\\
0.11	30.66\\
0.12	27.73\\
0.13	25.73\\
0.14	23.69\\
0.15	20.97\\
0.16	17.75\\
0.17	16.22\\
0.18	14.23\\
0.19	13.24\\
0.2	12.34\\
};
\addlegendentry{CGF};

\addplot [color=mycolor4,line width=1.5pt]
  table[row sep=crcr]{%
0.01	96.78\\
0.02	90.22\\
0.03	83.67\\
0.04	78.37\\
0.05	73.28\\
0.06	69.05\\
0.07	64.71\\
0.08	60.41\\
0.09	56.08\\
0.1	52.5\\
0.11	48.79\\
0.12	44.88\\
0.13	41.44\\
0.14	38.94\\
0.15	36.4\\
0.16	34.49\\
0.17	33.12\\
0.18	30.42\\
0.19	28.97\\
0.2	26.94\\
};
\addlegendentry{SHOT};

\addplot [color=mycolor5,line width=1.5pt]
  table[row sep=crcr]{%
0.01	95.83\\
0.02	87.89\\
0.03	79.14\\
0.04	67.64\\
0.05	57.34\\
0.06	48.53\\
0.07	41.97\\
0.08	36.03\\
0.09	31.33\\
0.1	26.96\\
0.11	24.02\\
0.12	20.55\\
0.13	17.91\\
0.14	16.57\\
0.15	13.95\\
0.16	12.2\\
0.17	10.69\\
0.18	9.39\\
0.19	8.33\\
0.2	7.73\\
};
\addlegendentry{3DMatch};

\addplot [color=mycolor6,line width=1.5pt]
  table[row sep=crcr]{%
0.01	93.12\\
0.02	83.23\\
0.03	73.7\\
0.04	63.79\\
0.05	54.33\\
0.06	47.39\\
0.07	41.67\\
0.08	37.32\\
0.09	32.64\\
0.1	30.25\\
0.11	26.13\\
0.12	23.24\\
0.13	20.99\\
0.14	19.18\\
0.15	16.26\\
0.16	14.34\\
0.17	12.38\\
0.18	10.6\\
0.19	9.48\\
0.2	8.67\\
};
\addlegendentry{FPFH};

\end{axis}
\end{tikzpicture}%
}
	\caption{\textbf{Recall in relation to inlier ratio threshold}. Recall of 3DSmoothNet on the \textit{3DMatch} data set remains high even when the inlier threshold ratio is increased.}
	\label{fig:threshold_results}
\end{figure}
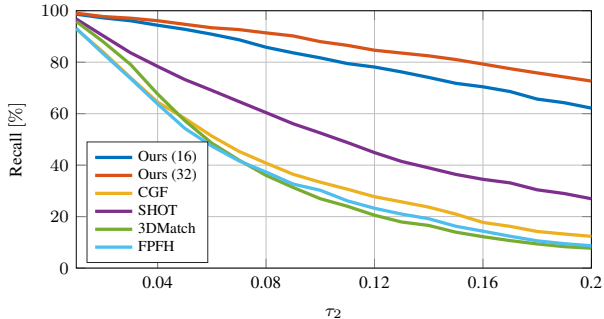
\vspace{-2ex}
\paragraph{Comparison to state-of-the-art}
Results of experimental evaluation on the \textit{3DMatch} data set are summarized in Tab.~\ref{tab:3DMatchResults}~(left) and two hard cases are shown in Fig.~\ref{fig:registarion_examples}. 
\textit{Ours (16)} and \textit{Ours (32)} achieve an average recall of $92.8\%$ and $94.7\%$, respectively, which is close to solving the \textit{3DMatch} data set. 3DSmoothNet outperforms all state-of-the-art 3D local feature descriptors with a significant margin on all scenes. Remarkably, \textit{Ours (16)} improves average recall over all scenes by almost $20$ percent points with only $16$ output dimensions compared to $512$ dimensions of PPF-FoldNet and $352$ of SHOT. Furthermore, \textit{Ours (16)} and \textit{Ours (32)} show a much smaller recall standard deviation (STD), which indicates robustness of 3DSmoothNet to scene changes and hints at good generalization ability. 
The inlier ratio threshold $\tau_2 = 0.05$ as chosen by~\cite{deng2018ppfnet} results in $\approx 55\text{k}$ iterations to find at least 3 correspondences (with $99.9 \%$ probability) with the common RANSAC approach. Increasing the inlier ratio to $\tau_2 = 0.2$ would decrease RANSAC iterations significantly to $\approx 850$, which would speed up processing massively.
We thus evaluate how gradually increasing the inlier ratio changes performance of 3DSmoothNet in comparison to all other tested approaches (Fig.~\ref{fig:threshold_results}). While the average recall of all other methods drops below $30\%$ for $\tau_2 = 0.2$, recall of \textit{Ours (16)} (blue) and \textit{Ours (32)} (orange) remains high at $62\%$ and $72\%$, respectively. This indicates that any descriptor-based point cloud registration pipeline can be made more efficient by just replacing the existing descriptor with our 3DSmoothNet.
\begin{table}[!t]
	\setlength{\tabcolsep}{4pt}
	\centering
	\begin{tabular}{l|d{2.1}d{2.1}|d{2.1}d{2.1}}
		\hline
		& \multicolumn{4}{c}{\textit{3DMatch} data set}\\
		\hline 
		& \multicolumn{2}{c|}{Original}  & \multicolumn{2}{c}{Rotated}\\
		& \multicolumn{1}{c}{Average}&\multicolumn{1}{c|}{STD}&\multicolumn{1}{c}{Average}&\multicolumn{1}{c}{STD}\\
		\hline\hline
		FPFH~\cite{rusu2009FPFH} & 54.3 & 11.8 & 54.8 & 12.1 \\
		SHOT~\cite{tombari2010SHOT}& 73.3 & 7.7 & 73.3 & 7.6  \\
		3DMatch~\cite{zeng20163dmatch}\footnotemark  & 57.3 & 7.8  & 3.6 & 1.7 \\
		CGF~\cite{khoury2017CGF} & 58.2 & 14.2 & 58.5 & 14.0 \\
		PPFNet~\cite{deng2018ppfnet}& 62.3 & 11.5 & 0.3 & 0.5 \\
		PPF-FoldNet~\cite{Deng2018PPFFoldNetUL} & 71.8 & 9.9 & 73.1 & 11.1 \\
		Ours ($16$ dim) & 92.8 & 3.4 & 93.0 & 3.2 \\
		Ours ($32$ dim) &\multicolumn{1}{B{.}{.}{-1}}{94.7} & 2.7 &\multicolumn{1}{B{.}{.}{-1}}{94.9} & 2.5 \\
		\hline
	\end{tabular}
	\caption{Results on the \textit{3DMatch} and \textit{3DRotatedMatch} data sets. We report average recall in percent over all scenes along with the standard deviation (STD) per method. Best performance is shown in bold. Note that results of non-rotation invariant methods naturally drop to zero for the rotated case (right column). See detailed results per scene in the Supplementary material.}
	\label{tab:3DMatchResults}
\end{table}
\vspace{-2ex}
\paragraph{Rotation invariance}
We take a similar approach as~\cite{Deng2018PPFFoldNetUL} to validate rotation invariance of 3DSmoothNet by rotating all fragments of \textit{3DMatch} data set (we name it \textit{3DRotatedMatch}) around all three axis and evaluating the performance of the selected descriptors on these rotated versions. Individual rotation angles are sampled arbitrarily between $[0, 2\pi]$ and the same indices of points for evaluation are used as in the previous section. Results of \textit{Ours (16)} and \textit{Ours (32)} remain basically unchanged (Tab~\ref{tab:3DMatchResults} (right)) compared to the non-rotated variant (Tab~\ref{tab:3DMatchResults} (left)), which confirms rotation invariance of 3DSmoothNet (due to estimating \acs{lrf}). Because performance of all other rotation invariant descriptors~\cite{rusu2009FPFH,tombari2010SHOT,khoury2017CGF,Deng2018PPFFoldNetUL} remains mainly identical, too, 3DSmoothNet again outperforms all state-of-the-art methods by more than $20$ percent points. 
\vspace{-2ex}
\paragraph{Ablation study} To get a better understanding of the reasons for the very good performance of 3DSmoothNet, we analyze the contribution of individual modules with an ablation study on \textit{3DMatch} and \textit{3DRotatedMatch} data sets. Along with the original 3DSmoothNet, we consider versions without \acs{sdv} (we use a simple binary occupancy grid), without \acs{lrf} and finally without both, \acs{lrf} and \acs{sdv}. All networks are trained using the same parameters and for the same number of epochs. Results of this ablation study are summarized in Tab.~\ref{tab:ablation_study}. 
\footnotetext{Using the precomputed feature descriptors provided by the authors. For more results see the Supplementary material.}
It turns out that the version without \acs{lrf} performs best on \textit{3DMatch} because most fragments are already oriented in the same way and the original data set version is tailored for descriptors that are not rotation invariant. Inferior performance of the full pipeline on this data set is most likely due to a few wrongly estimated \acs{lrf}, which reduces performance on already oriented data sets (but allows generalizing to the more realistic, rotated cases). Unsurprisingly, 3DSmoothNet without \acs{lrf} fails on \textit{3DRotatedMatch} because the network cannot learn rotation invariance from the data. A significant performance gain of up to more than $9$ percent points can be attributed to using an a \acs{sdv} voxel grid instead of the traditional binary occupancy grid. 
\begin{table}[!th]
	\setlength{\tabcolsep}{4pt}
	\centering
	\resizebox{\linewidth}{!}{\begin{tabular}{l|d{2.1}d{2.1}|d{2.1}d{2.1}}
			\hline
			& \multicolumn{4}{c}{\textit{3DMatch} data set}\\
			\hline 
			& \multicolumn{2}{c|}{Original}  & \multicolumn{2}{c}{Rotated}\\
			& \multicolumn{1}{c}{$\tau_2 = 0.05$}&  \multicolumn{1}{c|}{$\tau_2 = 0.2$} &  \multicolumn{1}{c}{$\tau_2= 0.05$} &  \multicolumn{1}{c}{$\tau_2 = 0.2$}\\
			\hline\hline
			All together &94.7 &72.7 & \multicolumn{1}{B{.}{.}{-1}}{94.9} & \multicolumn{1}{B{.}{.}{-1}}{72.8} \\
			W/o \acs{sdv} & 92.5 & 63.5  & 92.5 & 63.6  \\
			W/o \acs{lrf} & \multicolumn{1}{B{.}{.}{-1}}{96.3} & \multicolumn{1}{B{.}{.}{-1}|}{81.6}  & 11.6 & 2.7  \\
			W/o \acs{sdv} \& \acs{lrf}  & 95.6 & 78.6 & 9.7 & 2.1  \\
			\hline
	\end{tabular}}
	\caption{Ablation study of 3DSmoothNet on \textit{3DMatch} and \textit{3DRotatedMatch} data sets. We report average recall over all overlapping fragment pairs. Best performance is shown in bold.}
	\label{tab:ablation_study}
\end{table}
\begin{table}
	\setlength{\tabcolsep}{4pt}
	\centering
	\resizebox{\linewidth}{!}{\begin{tabular}{l|d{2.1}d{2.1}d{2.1}d{2.1}|d{2.1}}
			\hline
			&\multicolumn{2}{c}{Gazebo} & \multicolumn{2}{c|}{Wood} &  \\
			
			& \multicolumn{1}{c}{Sum.} & \multicolumn{1}{c}{Wint.} & \multicolumn{1}{c}{Aut.} & \multicolumn{1}{c|}{Sum.} & \multicolumn{1}{c}{Average} \\
			\hline\hline
			FPFH~\cite{rusu2009FPFH} & 38.6 & 14.2 & 14.8 & 20.8 & 22.1  \\
			SHOT~\cite{tombari2010SHOT} & 73.9 & 45.7 & 60.9 & 64.0 & 61.1 \\ 
			3DMatch~\cite{zeng20163dmatch} & 22.8 & 8.3 & 13.9 & 22.4 & 16.9  \\
			CGF~\cite{khoury2017CGF} & 37.5 & 13.8 & 10.4 & 19.2 & 20.2  \\
			Ours ( $16$  dim) & 76.1 & 47.7 & 31.3 & 37.6 & 48.2  \\
			Ours ( $32$  dim) & \multicolumn{1}{B{.}{.}{-1}}{91.3} & \multicolumn{1}{B{.}{.}{-1}}{84.1} & \multicolumn{1}{B{.}{.}{-1}}{67.8} & \multicolumn{1}{B{.}{.}{-1}|}{72.8} & \multicolumn{1}{B{.}{.}{-1}}{79.0}  \\
			\hline
	\end{tabular}}
	\caption{Results on the \textit{ETH}data set. We report average recall in percent per scene as well as across the whole data set.}
	\label{tab:ETH_Dataset}
\end{table}
\subsection{Generalizability across modalities and scenes}
We evaluate how 3DSmoothNet generalizes to outdoor scenes obtained using a laser scanner (Fig.~\ref{fig:teaser_photo}). To this end, we use models \textit{Ours (16)} and \textit{Ours (32)} trained on \textit{3DMatch} (RGB-D images of indoor scenes) and test on four outdoor laser scan data sets \textit{Gazebo-Summer, Gazebo-Winter, Wood-Autumn} and \textit{Wood-Summer} that are part of the \textit{ETH} data set~\cite{Pomerleau2012_ETHdataset}.
All acquisitions contain several partially overlapping scans of sparse and dense vegetation (e.g., trees and bushes). Accurate ground-truth transformation matrices are available through extrinsic measurements of the scanner position with a total-station. 
We start our evaluation by down-sampling the laser scans using a voxel grid filter of size $0.02\text{m}$. We randomly sample $5000$ points in each point cloud and follow the same evaluation procedure as in Sec~\ref{sec:3DMatch_data set}, again considering only point clouds with more than $30\%$ overlap. More details on sampling of the feature points and computation of the point cloud overlaps are available in the supplementary material. Due to the lower resolution of the point clouds, we now use a larger value of $W = 1~\text{m}$ for the \acs{sdv} voxel grid (consequently the radius for the descriptors based on the spherical neighborhood is also increased). A voxel grid with an edge equal to $1.5~\text{m}$ is used for 3DMatch because of memory restrictions. Results on the \textit{ETH} data set are reported in Tab~\ref{tab:ETH_Dataset}.
3DSmoothNet achieves best performance on average (right column), \textit{Ours (32)} with $79.0\%$ average recall clearly outperforming \textit{Ours (16)} with $48.2\%$ due to its larger output dimension. \textit{Ours (32)} beats runner-up (unsupervised) SHOT by more than $15$ percent points whereas all state-of-the-art methods stay significantly below $30\%$. In fact, \textit{Ours (32)} applied to outdoor laser scans still outperforms all competitors that are trained \textit{and} tested on the \textit{3DMatch} data set (cf. Tab.~\ref{tab:ETH_Dataset} with Tab.~\ref{tab:3DMatchResults}).  
\begin{table}[!t]
	\setlength{\tabcolsep}{4pt}
	\centering
	\resizebox{\linewidth}{!}{\begin{tabular}{l|cccc}
			\hline
			& Input prep. & Inference & NN search &  Total\\
			& [ms] &  [ms]  &  [ms]  &   [ms] \\
			\hline
			\hline
			3DMatch  &$0.5$ &$3.7$ &$0.8$ & $5.0$\\
			3DSmoothNet & $4.2$ & $0.3$ & $0.1$ & $4.6$\\
			\hline
	\end{tabular}}
	\caption{Average run-time per feature-point on test fragments of \textit{3DMatch} data set.}
	\label{tab:timing}
\end{table}
\vspace{-3ex}
\subsection{Computation time}
We compare average run-time of our approach per interest point on 3DMatch test fragments to~\cite{zeng20163dmatch} in Tab.~\ref{tab:timing} (ran on the same PC with Intel Xeon E5-1650, 32 GB of ram and NVIDIA GeForce GTX1080). Note that input preparation (Input prep.) and inference of~\cite{zeng20163dmatch} are processed on the GPU, while our approach does input preparation on CPU in its current state. For both methods, we run nearest neighbor correspondence search on the CPU. Naturally, input preparation of 3DSmoothNet on the CPU takes considerably longer (4.2 ms versus 0.5 ms), but still the overall computation time is slightly shorter (4.6 ms versus 5.0 ms). Main drivers for performance are inference (0.3 ms versus 3.7 ms) and nearest neighbor correspondence search (0.1 ms versus 0.8 ms). This indicates that it is worth investing computational resources into custom-tailored data preparation because it significantly speeds up all later tasks. The bigger gap between Ours(16 dim) and Ours(32 dim), is a result of the lower capacity and hence lower descriptiveness of the 16-dimensional descriptor, which becomes more apparent on the harder \textit{ETH} data set, but can also be seen in additional experiments in supplementary material. Supplementary material also contains additional experiments, which show the invariance of the proposed descriptor to changes in point cloud density. 
%

\section{Conclusions}\label{sec:conclusions}

We have presented 3DSmoothNet, a deep learning approach with fully convolutional layers for 3D point cloud matching that outperforms all state-of-the-art by more than $20$ percent points. It allows very efficient correspondence search due to low output dimensions (16 or 32), and a model trained on indoor RGB-D scenes generalizes well to terrestrial laser scans of outdoor vegetation. Our method is rotation invariant and achieves $94.9\%$ average recall on the \textit{3DMatch} benchmark data set, which is close to solving it. To the best of our knowledge, this is the first learned, universal point cloud matching method that allows transferring trained models between modalities. It takes our field one step closer to the utopian vision of a single trained model that can be used for matching any kind of point cloud regardless of scene content or sensor.

{\small
\bibliographystyle{ieee_fullname}
\bibliography{cvpr2019}
}
\newpage
\section{Supplementary Material}
\begin{figure*}[!ht]
	\begin{center}
		\includegraphics[width=0.95\linewidth]{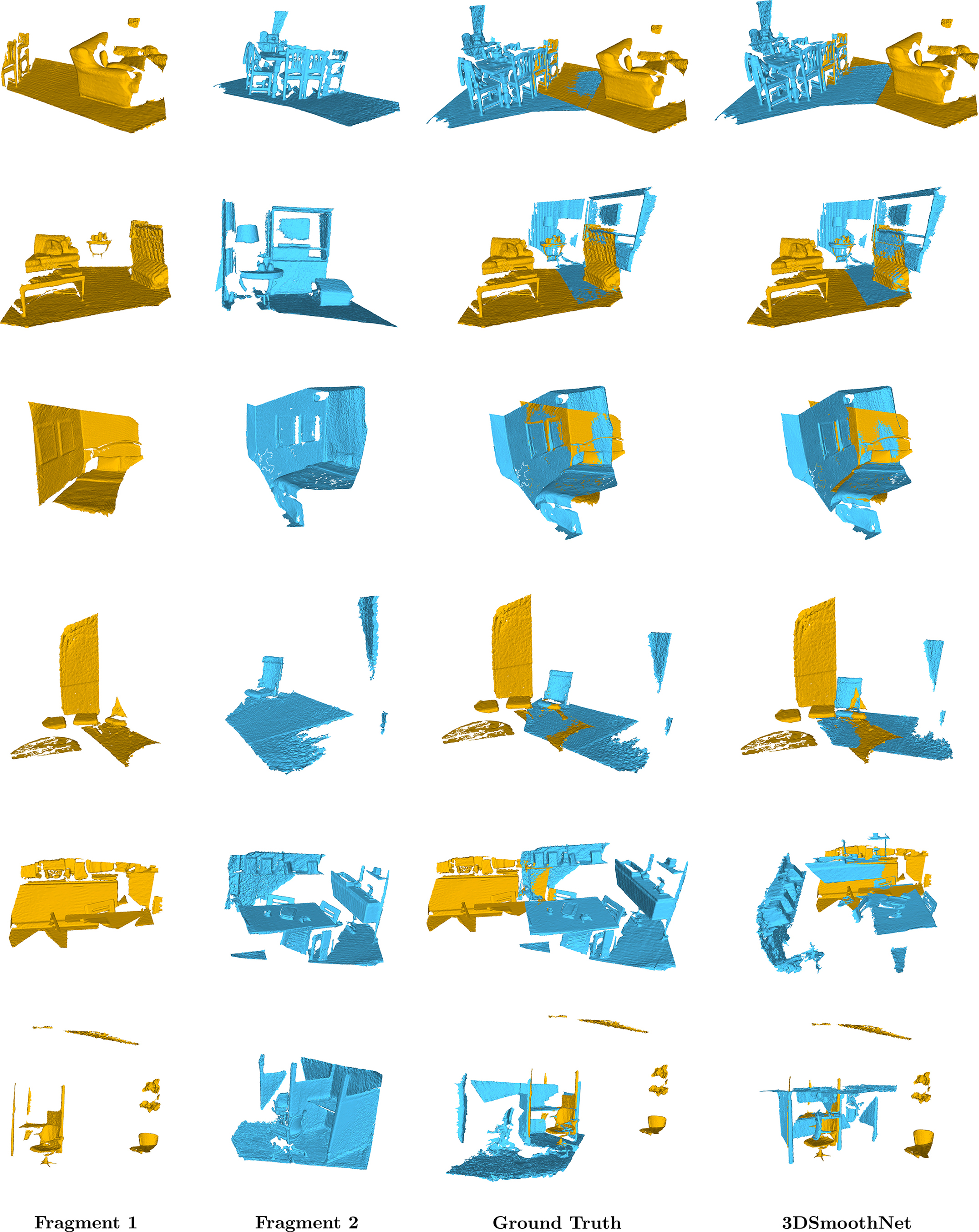}
	\end{center}
	\caption{\textbf{Additional qualitative results of 3DSmoothNet on the \textit{3DMatch} data set}. First three rows show hard examples for which the 3DSmoothNet succeeds, whereas the last three rows show some of the failure cases. 3DMatch and CGF fail for all these examples.}
	\label{fig:Visualization_3DMatch}
\end{figure*}

\begin{figure*}[!ht]
	\begin{center}
		\includegraphics[width=0.95\linewidth]{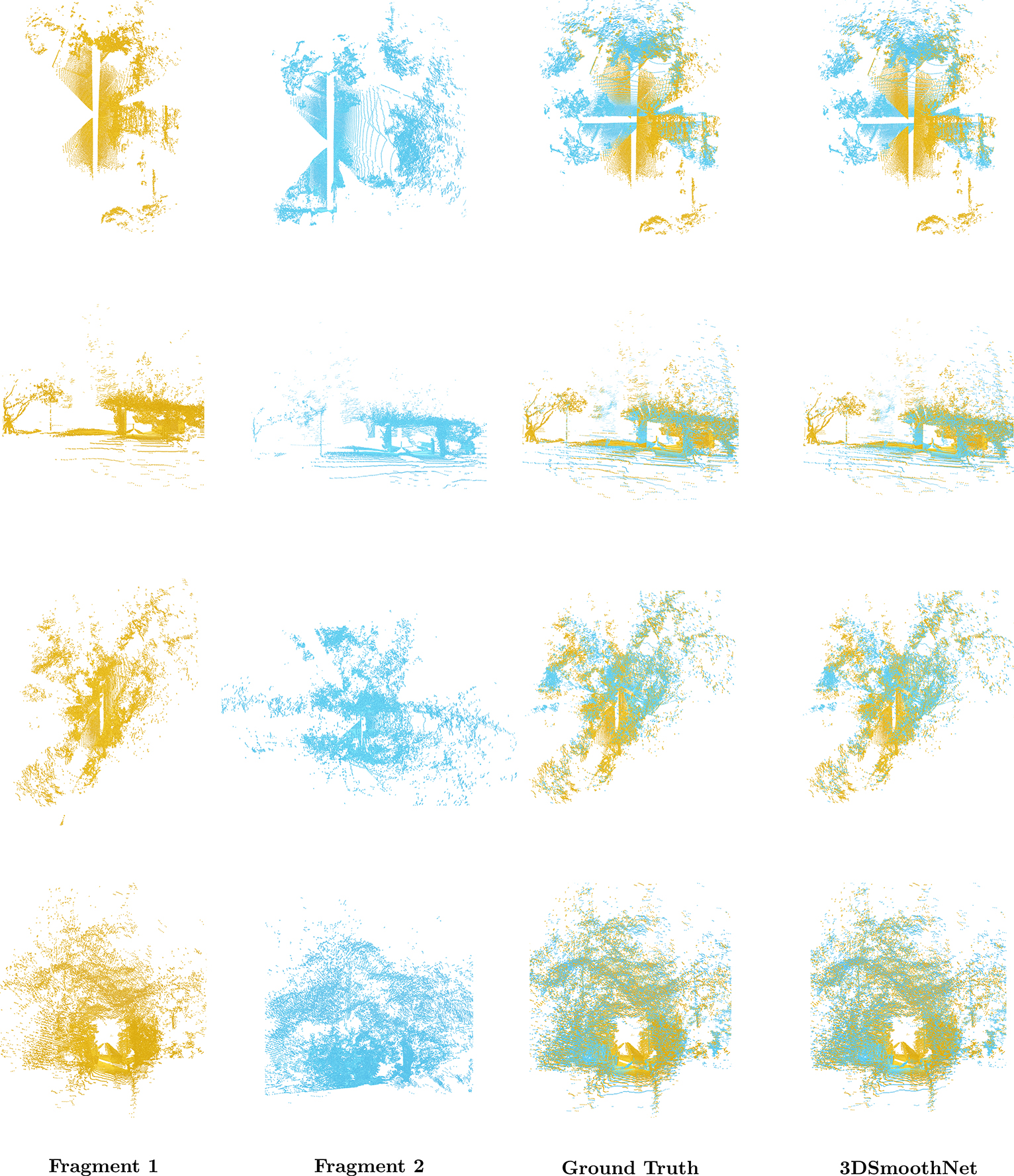}
	\end{center}
	\caption{\textbf{Qualitative results of the 3DSmoothNet on the \textit{ETH} data set}. 3DSmoothNet trainined only on the indoor reconstructions from RGB-D images can generalize to outdoor natural scenes, which consist of high level of noise and predominantly unstructured vegetation. The data set is made even harder by the introduced dynamic between the epochs (e.g. walking persons) }
	\label{fig:Visualization_ETH}
\end{figure*}

In this supplementary material we provide additional information about the evaluation experiments (Sec.~\ref{sec:evaluation_metric},~\ref{sec:base_param} and~\ref{sec:preprocessing}) along with the detailed per-scene results (Sec.~\ref{sec:results}) and some further visualizations (Fig.~\ref{fig:Visualization_3DMatch} and~\ref{fig:Visualization_ETH}). The source code and all the data needed for comparison are publicly available at \url{https://github.com/zgojcic/3DSmoothNet}.
\vspace{-1ex}
\subsection{Evaluation metric} \label{sec:evaluation_metric} This section provides a detailed explanation of the evaluation metric adopted from~\cite{deng2018ppfnet} and used for all evaluation experiments throughout the paper. 

Consider two point cloud fragments $\mathcal{P}$ and $\mathcal{Q}$, which have more than $30\%$ overlap under ground-truth alignment. Furthermore, let all such pairs form a set of fragment pairs $\mathcal{F} = \{(\mathcal{P},\mathcal{Q})\}$. For each fragment pair the set of correspondences obtained in the feature space is then defined as 
\begin{equation}
\begin{split}
\mathcal{C} = \{ \{\mathbf{p}_i \in \mathcal{P},\mathbf{q}_j \in \mathcal{Q} \}, f(\mathbf{p}_i) = \text{nn}(f(\mathbf{q}_j),f(\mathcal{P})) \land \\
 f(\mathbf{q}_j) = \text{nn}(f(\mathbf{p}_i),f(\mathcal{Q}))  \} 
\end{split}
\end{equation}
where $f(\mathbf{p})$ denotes a non-linear function that maps the feature point $\mathbf{p}$ to its local feature descriptor and $\text{nn()}$ denotes the nearest neighbor search based on the $l2$ distance. Finally, the quality of the correspondences in terms of average recall $R$ per scene is computed as
\begin{equation}
\resizebox{\linewidth}{!} 
{
$R = \frac{1}{|\mathcal{F}|} \sum\limits_{f=1}^{|\mathcal{F}|}  \mathbbm{1}\Bigl(\bigl[\frac{1}{|\mathcal{C}_f|}\sum\limits_{i,j \in \mathcal{C}_s} \mathbbm{1}\bigl( ||\mathbf{p}_i - T_f(\mathbf{q}_j) ||_2  < \tau_1 \bigr)\bigr]  > \tau_2 \Bigr) $
}
\end{equation}
where $T_f$ denotes the ground-truth transformation alignment of the fragment pair $f \in \mathcal{F}$. $\tau_1$ is the threshold on the Euclidean distance between the correspondence pair $(i,j)$ found in the feature space and $\tau_2$ is a threshold on the inlier ratio of the correspondences~\cite{deng2018ppfnet}. Following \cite{deng2018ppfnet} we set $\tau_1=0.1\text{m}$ and $\tau_2=0.05$ for both, the \textit{3DMatch}~\cite{zeng20163dmatch} as well as the \textit{ETH}~\cite{Pomerleau2012_ETHdataset} data set. The evaluation metric is based on the theoretical analysis of the number of iterations $k$ needed by RANSAC~\cite{fischer1981RANSAC} to find at least $n=3$ corresponding points with the probability of success $p=99.9\%$. Considering, $\tau_2 = 0.05$ and the relation 
\begin{equation}
k = \frac{\log(1-p)}{\log(1-\tau_2^n)},
\end{equation}
the number of iterations equals $k\approx 55000$ and can be greatly reduced if the number of inliers $\tau_2$ can be increased (e.g. $k = 860$ if $\tau_2 = 0.2$). 
\begin{table}[!t]
    \renewcommand{\arraystretch}{1.1}
    \setlength{\tabcolsep}{4pt}
	\centering
		\resizebox{\columnwidth}{!}{
    \begin{tabular}{l|cccc}
			\hline
			Method & Parameter & \textit{3Dmatch} data set & \textit{ETH} data set\\
			\hline\hline
			\multirow{2}{*}{FPFH~\cite{rusu2009FPFH}} & $r_f [\text{m}]$ &  $0.093$ & $0.310$ \\
			&  $r_n [\text{m}]$ & $0.093$ & $0.310$ \\
			\hline
			\multirow{2}{*}{SHOT~\cite{tombari2010SHOT}} & $r_f [\text{m}]$ &  $0.186$ & $0.620$\\
			&  $r_n [\text{m}]$ & $0.093$ & $0.310$\\
	        \hline
						\multirow{2}{*}{3DMatch~\cite{zeng20163dmatch}} & $W [\text{m}]$ &  $0.300$ & $1.500$\footnotemark\\
			&  $n_{voxels} $ & $30^3$ & $30^3$\\
			\hline
			\multirow{2}{*}{CGF~\cite{khoury2017CGF}} & $r_f [\text{m}]$ &  $0.186$ & $0.620$\\
			&  $r_n [\text{m}]$ & $0.093$ & $0.310$\\
			&  $r_\text{min} [\text{m}]$\footnotemark & $0.015$ & $0.05$\\
			\hline
						\multirow{2}{*}{PPFNet~\cite{deng2018ppfnet}} & $k_n [\text{points}]$ &  $17$ & / \\
			&  $r_{f} [\text{m}]$ & $0.300$ & / \\
			\hline
						\multirow{2}{*}{PPF-FoldNet~\cite{Deng2018PPFFoldNetUL}} & $k_n [\text{points}]$ &  $17$ & / \\
			&  $r_{f} [\text{m}]$ & $0.300$ & / \\
			\hline
	\end{tabular}}
	\caption{Parameters used for the state-of-the-art methods in the evaluation experiments.}
	\label{tab:parameters}
	\vspace{-1ex}
\end{table}
\footnotetext[3]{Larger voxel grid width used due to the memory restrictions.}
\footnotetext{Used to avoid the excessive binning near the center, see~\cite{khoury2017CGF}}
\begin{table*}[!t]
	\centering
	\resizebox{0.98\textwidth}{!}{\begin{tabular}{l|cccccccccccccc}
			\hline
			&FPFH~\cite{rusu2009FPFH}&SHOT~\cite{tombari2010SHOT}&3DMatch~\cite{zeng20163dmatch}&CGF~\cite{khoury2017CGF}&PPFNet~\cite{deng2018ppfnet}&PPF-FoldNet~\cite{Deng2018PPFFoldNetUL}&Ours &Ours\\
			&($33$ dim)&($352$ dim)&($512$ dim)&($32$ dim)&($64$ dim)&($512$ dim)&($16$ dim) &($32$ dim) \\
			\hline\hline
            Kitchen&$43.1$&$74.3$&$58.3$&$60.3$&$89.7$&$78.7$&$93.1$&$\mathbf{97.0}$&\\
            Home 1 &$66.7$&$80.1$&$72.4$&$71.1$&$55.8$&$76.3$&$93.6$&$\mathbf{95.5}$&\\
            Home 2 &$56.3$&$70.7$&$61.5$&$56.7$&$59.1$&$61.5$&$86.5$&$\mathbf{89.4}$&\\
            Hotel 1&$60.6$&$77.4$&$54.9$&$57.1$&$58.0$&$68.1$&$95.6$&$\mathbf{96.5}$&\\
            Hotel 2&$56.7$&$72.1$&$48.1$&$53.8$&$57.7$&$71.2$&$90.4$&$\mathbf{93.3}$&\\
            Hotel 3&$70.4$&$85.2$&$61.1$&$83.3$&$61.1$&$94.4$&$\mathbf{98.2}$&$\mathbf{98.2}$&\\
            Study  &$39.4$&$64.0$&$51.7$&$37.7$&$53.4$&$62.0$&$92.8$&$\mathbf{94.5}$&\\
            MIT Lab&$41.6$&$62.3$&$50.7$&$45.5$&$63.6$&$62.3$&$92.2$&$\mathbf{93.5}$&\\
			\hline
			Average&$54.3$&$73.3$&$57.3$&$58.2$&$62.3$&$71.8$&$92.8$&$\mathbf{94.7}$&\\
			\hline
			STD&$11.8$&$7.7$&$7.8$&$14.2$&$11.5$&$9.9$&$3.4$&$2.7$&\\
			\hline
	\end{tabular}}
	\caption{\textbf{Detailed quantitative results on the \textit{3DMatch} dataset.} For each scene we report the average recall in percent over all overlapping fragment pairs. Best performance is shown in bold.}
	\label{tab:results_3DMatch_dataset}
	\vspace{-2ex}
\end{table*}
\subsection{Baseline Parameters} \label{sec:base_param}
In order to perform the comparison with the state-of-the-art methods, several parameters have to be set. To ensure a fair comparison we set all the parameters relative to our voxel grid width $W$ which we set as $W_\textit{3DMatch} = 0.3\text{m}$ and $W_\textit{ETH} = 1\text{m}$ for \textit{3DMatch} and \textit{ETH} data sets respectively. More specific, for the descriptors based on the spherical support we use a feature radius $r_f = \sqrt[3]{\frac{3}{4\pi}}W$ that yields a sphere with the same volume as our voxel grid and for all voxel-based descriptors we use the same voxel grid width $W$. For descriptors that require, along with the coordinates also the normal vectors, we use the point cloud library (PCL) built-in function for normal vector computation, using all the points in the spherical support with the radius $r_n = \frac{r_f}{2}$. Tab.~\ref{tab:parameters} provides all the parameters that were used for the evaluation. If some parameters are not listed in Tab~\ref{tab:parameters} we use the original values set by the authors. For the handcrafted descriptors, FPFH~\cite{rusu2009FPFH} and SHOT~\cite{tombari2010SHOT} we use the implementation provided by the original authors as a part of the  PCL\footnote{https://github.com/PointCloudLibrary/pcl}. We use the PCL version $1.8.1$ x$64$ on Windows $10$ and use the parallel programming implementations (omp) of both descriptors. For 3DMatch~\cite{zeng20163dmatch} we use the implementation provided by the authors\footnote{https://github.com/andyzeng/3dmatch-toolbox} on Ubuntu 16.04 in combination with the CUDA $8.0$ and cuDNN $5.1$. Finally, for CGF~\cite{khoury2017CGF} we use the implementation provided by the authors\footnote{https://github.com/marckhoury/CGF} on a PC running Windows $10$. 
Note that we report the results of PPFNet~\cite{deng2018ppfnet} and PPF-FoldNet~\cite{Deng2018PPFFoldNetUL} as reported by the authors in the original papers, because the source code is not publicly available. Nevertheless, for the sake of completeness we report the feature radius $r_f$ and the k-nearest neighbors $k_n$ used for the normal vector computation, which were used by the authors in the original works. For the \textit{3DRotatedMatch} and \textit{3DSparseMatch} data sets we use the same parameters as for the\textit{3DMatch} data set.
\vspace{-2ex}
\paragraph{Performance of the 3DMatch descriptor}
The authors of the 3DMatch descriptor provide along with the source code and the trained model also the precomputed truncated distance function (TDF) representation and inferred descriptors for the \textit{3DMatch} data set. We use this descriptors directly for all evaluations on the original \textit{3DMatch} data set. For the evaluations on the \textit{3DRotatedMatch}, \textit{3DSparseMatch} and \textit{ETH} data sets we use their source code in combination with the pretrained model to infer the descriptors. When analyzing the \textit{3DSparseMatch} data set results, we noticed a discrepancy. The descriptors inferred by us achieve better performance than the provided ones. We analyzed this further and determined that the TDF representation (i.e. the input to the CNN) is identical and the difference stems from the inference using their provided weights. In the paper this is marked by a footnote in the results section. For the sake of consistency, we report in this Supplementary material all results for \textit{3DMatch} data set using the precoumpted descriptors and the results on all other data set using the descriptors inferred by us.
\subsection{Preprocessing of the benchmark data sets} \label{sec:preprocessing}
\paragraph{\textit{\textbf{3DMatch}} data set}
The authors of $\textit{3DMatch}$ data set provide along with the point cloud fragments and the ground-truth transformation parameters also the indices of the interest points and the ground-truth overlap for all fragments. To make the results comparable to previous works, we use these indices and overlap information for all descriptors and perform no preprocessing of the data.
\vspace{-2ex}
\paragraph{\textit{\textbf{3DSparseMatch}} data set}
In order to test the robustness of our approach to variations in point density we create a new data set, denoted as \textit{3DSparseMatch}, using the point cloud fragments from the \textit{3DMatch} data set. Specifically, we first extract the indices of the interest points provided by the authors of the \textit{3DMatch} data set and then randomly downsample the remaining points, keeping $50\%$, $25\%$ and $12.5\%$ of the points. We consider two scenarios in the evaluation. In the first scenario we use one of the fragments to be registered with the full and the other one with the reduced point cloud density (\textit{Mixed}), while in the second scenario we evaluate the descriptors on the fragments with the same level of sparsity (\textit{Both}).
\begin{table}[!b]
	\centering
	\setlength{\tabcolsep}{3pt}
	\resizebox{\linewidth}{!}{\begin{tabular}{l|cccccccccccc}
			\hline
			&	FPFH&	SHOT&	3DMatch&	CGF&	Ours &	Ours\\
			&	($33$ dim)&	($352$ dim)&	($512$ dim)&	($32$ dim)&	($16$ dim) &	($32$ dim) \\
			\hline\hline
			Kitchen&	$89$&	$154$&	$103$&	$125$&	$200$&	$\mathbf{274}$&	\\
			Home 1 &	$142$&	$206$&	$134$&	$156$&	$252$&	$\mathbf{324}$&	\\
			Home 2 &	$125$&	$182$&	$125$&	$142$&	$247$&	$\mathbf{318}$&	\\
			Hotel 1&	$86$&	$131$&	$73$&	$90$&	$192$&	$\mathbf{272}$&	\\
			Hotel 2&	$94$&	$124$&	$64$&	$94$&	$178$&	$\mathbf{238}$&	\\
			Hotel 3&	$119$&	$159$&	$64$&	$130$&	$210$&	$\mathbf{276}$&	\\
			Study  &	$56$&	$84$&	$64$&	$55$&	$130$&	$\mathbf{171}$&	\\
			MIT Lab&	$74$&	$121$&	$84$&	$78$&	$194$&	$\mathbf{246}$&	\\
			\hline
			Average&	$98$&	$145$&	$88$&	$108$&	$200$&	$\mathbf{264}$&	\\
			\hline
	\end{tabular}}
	\caption{\textbf{Average number of correct correspondences on \textit{3DMatch} data set.}  We report the average number of correct correspondences over all overlapping fragments of individual scenes.}
	\label{tab:number_matches}
\end{table}
\begin{table*}[!t]
	\centering
	\resizebox{0.98\textwidth}{!}{\begin{tabular}{l|cccccccccccccc}
			\hline
			&FPFH~\cite{rusu2009FPFH}&SHOT~\cite{tombari2010SHOT}&3DMatch~\cite{zeng20163dmatch}&CGF~\cite{khoury2017CGF}&PPFNet~\cite{deng2018ppfnet}&PPF-FoldNet~\cite{Deng2018PPFFoldNetUL}&Ours &Ours\\
			&($33$ dim)&($352$ dim)&($512$ dim)&($32$ dim)&($64$ dim)&($512$ dim)&($16$ dim) &($32$ dim) \\
			\hline\hline
			Kitchen&$43.5$&$74.1$&$2.4$&$60.5$&$0.2$&$78.9$&$93.3$&$\mathbf{97.2}$&\\
			Home 1 &$66.7$&$80.1$&$3.8$&$71.2$&$0.0$&$78.2$&$93.6$&$\mathbf{96.2}$&\\
			Home 2 &$56.3$&$70.2$&$5.3$&$57.2$&$1.4$&$64.4$&$87.0$&$\mathbf{90.9}$&\\
			Hotel 1&$62.4$&$77.0$&$1.8$&$57.2$&$0.4$&$67.7$&$95.6$&$\mathbf{96.5}$&\\
			Hotel 2&$56.7$&$72.1$&$6.7$&$53.8$&$0.0$&$69.2$&$91.4$&$\mathbf{92.3}$&\\
			Hotel 3&$72.2$&$85.2$&$1.9$&$83.3$&$0.0$&$96.3$&$\mathbf{98.2}$&$\mathbf{98.2}$&\\
			Study  &$39.7$&$65.1$&$2.7$&$38.7$&$0.0$&$62.7$&$93.2$&$\mathbf{94.5}$&\\
			MIT Lab&$41.6$&$62.3$&$3.9$&$45.5$&$0.0$&$67.5$&$92.2$&$\mathbf{93.5}$&\\
			\hline
			Average&$54.9$&$73.3$&$3.6$&$58.5$&$0.3$&$73.1$&$93.0$&$\mathbf{94.9}$&\\
			\hline
			STD&$12.2$&$7.6$&$1.7$&$14.0$&$0.5$&$11.1$&$3.2$&$2.5$&\\
			\hline
	\end{tabular}}
	\caption{\textbf{Detailed quantitative results on the \textit{3DRotatedMatch} data set.} For each scene we report the average recall in percent over all overlapping fragment pairs. Best performance is shown in bold.}
	\label{tab:results_3DRotatedMatch_dataset}
	\vspace{-1.5ex}
\end{table*}
\vspace{-1ex}
\paragraph{\textit{\textbf{ETH}} data set}
For the \textit{ETH} data set we use the point clouds and the ground-truth transformation parameters provided by the authors of the data set. We start by downsampling the point clouds using a voxel grid filter with the voxel size equal to $0.02\text{m}$. The authors of the data set also provide the ground-truth overlap information, but due to the downsampling step we opt to compute the overlap on our own as follows. Let $\mathbf{p}_i \in \mathcal{P}$ and $\mathbf{q}_i \in \mathcal{Q}$ denote points in the point clouds $\mathcal{P}$ and $\mathcal{Q}$, which are part of the same scene of \textit{ETH} data set, respectively. Given the ground-truth transformation $T_\mathcal{P}^\mathcal{Q}$ that aligns the point cloud $\mathcal{Q}$ with the point cloud $\mathcal{P}$, we compute the overlap $\psi_{\mathcal{P},\mathcal{Q}}$ relative to point cloud $\mathcal{P}$ as
\begin{equation}
    \psi_{\mathcal{P},\mathcal{Q}} = \frac{1}{|\mathcal{P}|} \sum\limits_{i=1}^{|\mathcal{P}|} \mathbbm{1}\bigl( ||\mathbf{p}_i - \text{nn}(\mathbf{p}_i,T_{\mathcal{P},\mathcal{Q}}(\mathcal{Q})  ||_2 < \tau_\psi \bigr)
\end{equation}
where nn denotes the nearest neighbor search based on the $l2$ distance in the Euclidean space and $\tau_\psi$ thresholds the distance between the nearest neighbors. In our evaluation experiments, we select  $\tau_\psi= 0.06\text{m}$, which equals three times the resolution of the point clouds after the voxel grid downsampling, and consider only the point cloud pairs for which both $\psi_{\mathcal{P},\mathcal{Q}}$ and $\psi_{\mathcal{Q},\mathcal{P}}$ are bigger than $0.3$.
Because no indices of the interest points are provided we randomly sample $5000$ interest points that have more than $10$ neighbor points in a sphere with a radius $r=0.5\text{m}$ in every point cloud. The condition of minimum ten neighbors close to the interest point is enforced in order to avoid the problems with the normal vector computation. 
\subsection{Detailed results}  \label{sec:results}
\paragraph{\textit{\textbf{3DMatch}} data set}
Detailed per scene results on the \textit{3DMatch} data set are reported in Tab.~\ref{tab:results_3DMatch_dataset}. \textit{Ours} (32 dim) consistently outperforms all state-of-the-art by a significant margin and achieves a recall higher than $89\%$ on all of the scenes. However, the difference between the performance of individual descriptors is somewhat masked by the selected low value of $\tau_2$, e.g. same average recall on Hotel 3 scene achieved by \textit{Ours} (16 dim) and \textit{Ours} (32 dim). Therefore, we additionally perform a more direct evaluation of the quality of found correspondences, by computing the average number of correct correspondences established by each individual descriptor (Tab~\ref{tab:number_matches}). Where the term correct correspondences, denotes the correspondences for which the distance between the points in the coordinate space after the ground-truth alignment is smaller than $0.1\text{m}$.  Results in Tab.~\ref{tab:number_matches} again show the dominant performance of the 3DSmoothNet compared to the other state-of-the-art but also highlight the difference between \textit{Ours} (32 dim) and \textit{Ours} (16 dim). Remarkably, \textit{Ours} (32 dim) can establish almost two times more correspondences than the closest competitor. 
\paragraph{\textit{\textbf{3DRotatedMatch}} data set}
We additionally report the detailed results on the \textit{3DRotatedMatch} data set in Tab~\ref{tab:results_3DRotatedMatch_dataset}. Again, 3DSmoothNet outperforms all other descriptor on all the scenes and maintains a similar performance as on the \textit{3DMatch} data set. As expected the performance of the rotational invariant descriptors~\cite{rusu2009FPFH,tombari2010SHOT,khoury2017CGF,Deng2018PPFFoldNetUL} is not affected by the rotations of the fragments, whereas the performance of the descriptors, which are not rotational invariant~\cite{zeng20163dmatch,deng2018ppfnet} drops to almost zero. This greatly reduces the applicability of such descriptors for general use, where one considers the point cloud, which are not represented in their canonical representation. 
\paragraph{\textit{\textbf{3DSparseMatch}} data set}
Tab~\ref{tab:3DSparseMatchResults} shows the results on the three different density levels ($50\%$, $25\%$ and $12,5\%$) of the \textit{3DSparseMatch} data set. Generally, all descriptors perform better when the point density of only one fragments is reduced, compared to when both fragments are downsampled. In both scenarios, the recall of our approach drops marginally by max $1$ percent point and remains more than 20 percent points above any other competing method. Therefore, 3DSmoothNet can be labeled as invariant to point density changes.

\begin{table}[!b]
    \setlength{\tabcolsep}{4pt}
	\centering
    \resizebox{\columnwidth}{!}{\begin{tabular}{l|d{2.1}d{2.1}d{2.1}|d{2.1}d{2.1}d{2.1}}
			\hline
			& \multicolumn{6}{c}{\textit{3DSparseMatch} data set}\\
			\hline 
			& \multicolumn{3}{c|}{\textit{Mixed}}  & \multicolumn{3}{c}{\textit{Both}}\\
			& \multicolumn{1}{c}{$50\%$}&\multicolumn{1}{c}{$25\%$}&\multicolumn{1}{c|}{$12.5\%$}&\multicolumn{1}{c}{$50\%$}&\multicolumn{1}{c}{$25\%$}&\multicolumn{1}{c}{$12.5\%$}\\
			\hline\hline
			FPFH~\cite{rusu2009FPFH} & 54.4 & 52.0 & 48.3 & 52.2 & 49.7 & 41.5 \\
			SHOT~\cite{tombari2010SHOT} & 71.1 & 69.8 & 69.8 & 70.8 & 68.4 & 66.4 \\
			3DMatch~\cite{zeng20163dmatch}  & 73.0 & 72.7 & 70.2 & 73.8 & 72.8 & 72.8 \\
			CGF~\cite{khoury2017CGF} & 54.2 & 49.0 & 37.5 & 50.3 & 38.3 & 24.4 \\
			Ours ($16$ dim) & 92.5 & 92.3 & 91.3 & 92.7 & 91.7 & 90.5 \\
			Ours ($32$ dim) &\multicolumn{1}{B{.}{.}{-1}}{95.0} & \multicolumn{1}{B{.}{.}{-1}}{94.5}&
			\multicolumn{1}{B{.}{.}{-1}|}{94.1}& \multicolumn{1}{B{.}{.}{-1}}{95.0} & \multicolumn{1}{B{.}{.}{-1}}{94.5}&
			\multicolumn{1}{B{.}{.}{-1}}{93.7} \\
			\hline
	\end{tabular}}
	\caption{Results on the \textit{3DSparseMatch} data set. 'Mixed' denotes Scenario 1 in which only one of the fragments was downsampled and 'Both' denotes that both fragments were downsampled. We report average recall in percent over all scenes. Best performance is shown in bold.}
	\label{tab:3DSparseMatchResults}
\end{table}

\end{document}